\documentclass[10pt,twocolumn,letterpaper]{article}

\usepackage{cvpr}              % To produce the CAMERA-READY version
\usepackage{graphicx}
\usepackage{amsmath}
\usepackage{amssymb}
\usepackage{booktabs}
\usepackage{multirow}
\usepackage[dvipsnames]{xcolor}

\usepackage{cellspace}
\newcommand\cincludegraphics[2][]{\raisebox{-0.3\height}{\includegraphics[#1]{#2}}}

\usepackage{tikz}
\def\checkmark{\tikz\fill[scale=0.4](0,.35) -- (.25,0) -- (1,.7) --(.25,.15) -- cycle;} 

\usepackage[pagebackref,breaklinks,colorlinks]{hyperref}

\usepackage[capitalize]{cleveref}
\crefname{section}{Sec.}{Secs.}
\Crefname{section}{Section}{Sections}
\Crefname{table}{Table}{Tables}
\crefname{table}{Tab.}{Tabs.}

\global\long\def\comma{\enspace\mbox{,}}%
\global\long\def\dotmath{\enspace\mbox{.}}
\global\long\def\severity{sev} % Jan, you have to decide whether you prefer "off" to "sev" !

\def\cvprPaperID{15} % *** Enter the CVPR Paper ID here
\def\confName{CVSports}
\def\confYear{2024}
\newcommand{\mysection}[1]{\vspace{2pt}\noindent\textbf{#1}}
\usepackage{xcolor, color, colortbl}         % colors
\definecolor{Highlight}{HTML}{39b54a}  % green

\usepackage{amssymb}% http://ctan.org/pkg/amssymb
\usepackage{pifont}% http://ctan.org/pkg/pifont
\usepackage{algorithm}
\usepackage{listings}
\usepackage{etoolbox}
\makeatletter
\AfterEndEnvironment{algorithm}{\let\@algcomment\relax}
\AtEndEnvironment{algorithm}{\kern2pt\hrule\relax\vskip3pt\@algcomment}
\let\@algcomment\relax
\newcommand\algcomment[1]{\def\@algcomment{\footnotesize#1}}
\renewcommand\fs@ruled{\def\@fs@cfont{\bfseries}\let\@fs@capt\floatc@ruled
  \def\@fs@pre{\hrule height.8pt depth0pt \kern2pt}%
  \def\@fs@post{}%
  \def\@fs@mid{\kern2pt\hrule\kern2pt}%
  \let\@fs@iftopcapt\iftrue}
\makeatother
\newcommand{\cmmnt}[1]{}

\usepackage[normalem]{ulem}

\usepackage{listings}
\usepackage{color}
\definecolor{codegreen}{rgb}{0,0.6,0}
\definecolor{codegray}{rgb}{0.5,0.5,0.5}
\definecolor{codepurple}{rgb}{0.58,0,0.82}
\definecolor{backcolour}{rgb}{1,1,1}
 
\lstdefinestyle{mystyle}{
    backgroundcolor=\color{backcolour},   
    commentstyle=\color{codegreen},
    keywordstyle=\color{magenta},
    numberstyle=\tiny\color{codegray},
    stringstyle=\color{codepurple},
    basicstyle=\footnotesize,
    breakatwhitespace=false,         
    breaklines=true,                 
    captionpos=b,                    
    keepspaces=true,                 
    numbers=left,                    
    numbersep=5pt,                  
    showspaces=false,                
    showstringspaces=false,
    showtabs=false,                  
    tabsize=2
}
 
\makeatletter

\newcommand{\Rmnum}[1]{\expandafter\@slowromancap\romannumeral #1@}
\makeatother
\begin{document}

% \title{MVAS: Multi-View Soccer Videos Understanding for Referees Assistance}
% \title{MVAS: a Multi-View Assistant System to Assist Referee in Fouls Decision Making}
% \title{Leveling Up Fair Play in Soccer: A Multi-View Assistant System for Accurate Referee Decision Making on Fouls}
% \title{MVAS: Multi-View Assistant System \\for Accurate Referee Decision Making in Soccer}
\title{X-VARS: Introducing Explainability in Football Refereeing\\with Multi-Modal Large Language Models}
% \title{VARS: Towards Automated Soccer Decision Making from Multiple Views}

\author{Jan Held$^{1}$%\\ {\small Université Libre de Bruxelles} \and
\quad
Hani Itani$^{2}$  %\\ {\small KAUST}  \and 
\quad
Anthony Cioppa$^{1}$  %\\ {\small University of Liège, KAUST}  \and
\quad
Silvio Giancola$^{2}$  %\\ {\small KAUST}  \and 
\\
Bernard Ghanem$^{2}$  %\\ {\small KAUST}  \and
\quad
Marc Van Droogenbroeck$^{1}$  %\\ {\small University of Liège}
\\ $^1$ {\small University of Liège}
\quad $^2$ {\small KAUST}
}
\maketitle

%%%%%%%%% ABSTRACT
\begin{abstract}
%
%
% GENERIC INTRO OF THE FIELD
%The integration of artificial intelligence in sports, particularly in soccer, has seen significant advancements in recent years. The ability to process and interpret live game footage using AI has opened new avenues for enhancing game analysis, strategy development, and captioning.
% CURRENT ISSUE IN THAT FIELD
%One of the primary issues in this domain is the lack of AI models that can effectively combine videos with advanced language understanding capabilities for visual question-answering, captioning, or ruled-based reasoning.
%
% GENERIC INTRO OF THE FIELD
The rapid advancement of artificial intelligence has led to significant improvements in automated decision-making. However, the increased performance of models often comes at the cost of explainability and transparency of their decision-making processes. 
% CURRENT ISSUE IN THAT FIELD
In this paper, we investigate the capabilities of large language models to explain decisions, using football refereeing as a testing ground, given its decision complexity and subjectivity.
% OUR SOLUTION
% \newline
We introduce the E\textbf{\underline{X}}plainable \textbf{\underline{V}}ideo \textbf{\underline{A}}ssistant \textbf{\underline{R}}eferee \textbf{\underline{S}}ystem, X-VARS, a multi-modal large language model designed for understanding football videos from the point of view of a referee. 
% MORE INFO ON THAT SOLUTION
X-VARS can perform a multitude of tasks, including video description, question answering, action recognition, and conducting meaningful conversations based on video content and in accordance with the Laws of the Game for football referees.
% MORE ON EXPERIMENTAL VALIDATION
% GENERIC CONCLUSION ON THE IMPACT
We validate X-VARS on our novel dataset, \emph{SoccerNet-XFoul}, which consists of more than $22$k video-question-answer triplets annotated by over $70$ experienced football referees. 
Our experiments and human study illustrate the impressive capabilities of X-VARS in interpreting complex football clips. 
Furthermore, we highlight the potential of X-VARS to reach human performance and support football referees in the future.
% \newline
We will provide code, model, dataset, and demo upon publication.
%in supplementary material.

\end{abstract}

%%%%%%%%% BODY TEXT
\section{Introduction}
\label{sec:intro}

\begin{figure}[t]
    \centering
    \includegraphics[width=\linewidth]{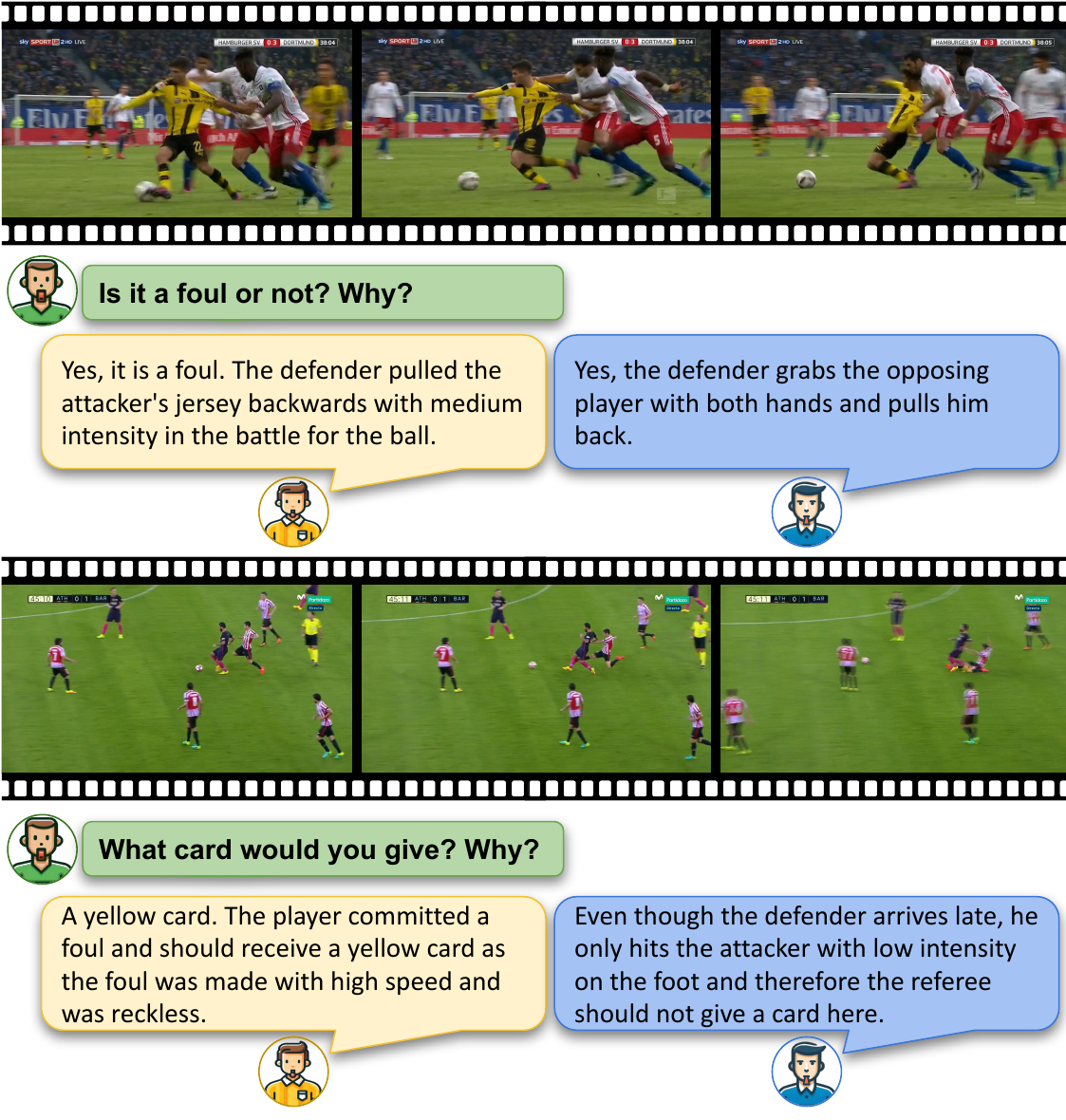}
    \caption{\textbf{SoccerNet-XFoul dataset.} 
Examples of annotations from two different referees for the same foul.
   The second example illustrates the complexity and subjectivity of refereeing decisions. %Two referees provide different interpretations of the same foul.
    }
    \label{fig:dataset_example}
\end{figure}

Recently, the field of Artificial Intelligence (AI) has witnessed remarkable progress in the development of large language models~\cite{Touvron2023LLaMA-arxiv, Brown2020Language-arxiv, Devlin2018BERT-arxiv, Hoffmann2022Training-arxiv, Chowdhery2022PaLM-arxiv}.
These models have acquired a strong language understanding, enabling them to tackle a broad range of linguistic tasks, ranging from text generation and conversation to zero-shot question answering.
The development has further progressed with multi-modal language models, going beyond the constraints of text-based inputs, but also including images, videos, and audio~\cite{Zhu2023MiniGPT4-arxiv, Maaz2023VideoChatGPT-arxiv, Ye2023mPLUGOwl2-arxiv, Alayrac2022Flamingo-arxiv}.
However, as the capabilities of these models continue to advance, the increase in model performance often comes at the cost of reduced explainability and transparency. 
This trend poses several challenges for users and developers who seek to understand the \textit{why} and \textit{how} behind a model's decision-making process. 
Explaining the reasoning process of AI models is particularly crucial in domains requiring high levels of trust, such as healthcare, autonomous driving, or sports.
In the context of football, referee decisions can significantly impact the financial future and existence of clubs, making it essential for AI models to transparently explain their decision-making process. Such transparency is key to building trust and facilitating the acceptance and integration of AI in sports.
For instance, during the Qatar World Cup 2022, FIFA employed an AI system for semi-automated offside detection~\cite{FIFA2023SemiAutomatedOffsideTechnology}.
To enhance the system's transparency and explainability, it generates a 3D representation of the game to allow referees and spectators to visually verify offside positions with undeniable clarity, bridging the gap between AI decision-making and human understanding.

In this paper, we explore the use of large language models to enhance transparency and explainability of automated referee decision-making. % through textual explanations. 
Football refereeing offers an ideal environment, as many of the decisions that referees make are subjective and reliant on each individual's interpretation of the rules of the game. 
Particularly, we introduce X-VARS, the first multi-modal large language model designed to explain football refereeing decisions.
X-VARS is based on a Vision Language Model (VLM)~\cite{Maaz2023VideoChatGPT-arxiv} that adapts the design of the LLaVa~\cite{Liu2023Visual-arxiv} model for spatio-temporal video modeling. 
We train X-VARS using a new training paradigm where we input the visual features and the multi-task predictions of our fine-tuned visual encoder CLIP ViT-L/14~\cite{Radford2021Learning} to the language model. 
We validate X-VARS on our novel \textit{SoccerNet-XFoul} dataset containing more than $22$k video-question-answer triplets about the most fundamental refereeing questions. More than $70$ professional referees annotated our dataset and provided, for each question, detailed explanations about their decisions.
X-VARS achieves state-of-the-art performance on the \textit{SoccerNet-MVFoul} dataset, and our human study demonstrates that X-VARS generates explanations for its decisions at a level comparable to human referees. 
Finally, X-VARS can analyze and understand complex football duels and provide accurate decision explanations, opening doors for future applications to support referees in their decision-making processes.

\mysection{Contributions.} We summarize our contributions as follows:
\textbf{(i)} We publicly release \textit{SoccerNet-XFoul}, a new multi-modal dataset containing more than $22$k video-question-answer triplets about refereeing questions. 
\textbf{(ii)} We introduce \textit{X-VARS}, a new vision language model that can perform multiple multi-modal tasks such as visual captioning, question-answering, video action recognition, and can generate explanations of its decisions on-par with human level.
\textbf{(iii)} We perform a thorough evaluation of our model, including analyses of our new training paradigm, the influence of the CLIP text predictions, and a human study that compares X-VARS to human referees.

\section{Related work}
\label{sec:related_work}

\mysection{Sports video understanding.}
The field of sports video understanding has gained a lot of interest lately~\cite{Thomas2017Computer}. It encompasses a wide range of tasks such as player segmentation, detection, and tracking~\cite{Cioppa2019ARTHuS, Maglo2022Efficient, Vandeghen2022SemiSupervised, Mansourian2023Multitask, Seweryn2024Improving-arxiv}, keypoint detection~\cite{Ludwig2023AllKeypoints}, summarizing~\cite{Gautam_2022, Midoglu_2024}, player re-identification~\cite{Somers2023Body, Mansourian2023Multitask}, action spotting in untrimmed videos~\cite{Cioppa2020AContextaware, Cioppa2021Camera,Hong2022Spotting,Soares2022Temporally, Soares2022Action-arxiv, Giancola2023Towards, Seweryn2023Survey-arxiv, Denize2023COMEDIAN-arxiv}, pass prediction and feasibility~\cite{ArbuesSanguesa2020Using, Honda2022Pass}, foul recognition~\cite{Held2023VARS, Fang2024Foul-arxiv} or dense video captioning for football broadcasts commentaries~\cite{Mkhallati2023SoccerNetCaption}. 
Such tasks can be formulated and solved by leveraging recent advances in deep learning for general video understanding. However, progress in sports video understanding heavily relies on the availability of sports-centric large-scale annotated datasets~\cite{Pappalardo2019Apublic,Yu2018Comprehensive,Scott2022SoccerTrack,Jiang2020SoccerDB,VanZandycke2022DeepSportradarv1, Istasse2023DeepSportradarv2, Midoglu2022MMSys}.
SoccerNet~\cite{Giancola2018SoccerNet,
Deliege2021SoccerNetv2,
Cioppa2022Scaling, 
Cioppa2022SoccerNetTracking, 
Giancola2022SoccerNet, 
Cioppa2023SoccerNetChallenge-arxiv} 
stands among the largest and most comprehensive dataset for video understanding in soccer. \emph{SoccerNet-MVFoul}~\cite{Held2023VARS} further extended \textit{SoccerNet} by proposing a novel multi-view football dataset designed for foul classification annotated by professional referees.
In this work, we further extend \emph{SoccerNet-MVFoul} into a visual-question-answering dataset focused on football refereeing questions, named \emph{SoccerNet-XFoul}.

\mysection{Vision language models.}
Natural Language Processing (NLP) has witnessed remarkable advancements with the emergence of open-source Large Language Models (LLMs)~\cite{Touvron2023LLaMA-arxiv, Brown2020Language-arxiv, Devlin2018BERT-arxiv, Hoffmann2022Training-arxiv, Chowdhery2022PaLM-arxiv, Ouyang2022Training-arxiv, Workshop2022BLOOM-arxiv}. These models have demonstrated exceptional capabilities in language understanding and generation tasks. LLMs have also served as the basis for the success of many vision-language models that are based on projecting the visual features of an image~\cite{Zhu2023MiniGPT4-arxiv, Chen2023ShareGPT4V-arxiv, Huang2023Language-arxiv, Xu2022BridgeTower-arxiv, Liu2023Visual-arxiv} or a video~\cite{Ye2023mPLUGOwl2-arxiv, Alayrac2022Flamingo-arxiv} encoder onto the input embedding space of an LLM. This idea and its variants allowed leveraging the power of LLMs for multi-modal understanding. 
In this work, we propose X-VARS, a vision language model for visual captioning, question-answering, action video recognition, and conducting meaningful conversations based on video content.

% a trainable module to bridge the gap between a frozen image encoder and a frozen LLM. It employs a set of learnable query vectors to extract visual features from the frozen image encoder. It constitutes an information bridge between the frozen image encoder and the frozen LLM, and feeds the most important visual feature to the LLM to output the desired text. 
% Based on this work, many researchers have published LLMs capable of accepting diverse inputs, such as images, videos , video and audio \cite{Zhang2023VideoLLaMA-arxiv}, and have further extended them to cover images, videos, and audio as inputs \cite{Chen2023XLLM-arxiv, Lyu2023MacawLLM-arxiv}. 
%  and LLava \cite{Liu2023Visual-arxiv} architecture and extend it to a multi-view video input.

\mysection{Explainability.} Recently, explainability in machine learning has received lots of attention, leading to the development of various techniques to demystify complex models. LIME (Local Interpretable Model-agnostic Explanations)\cite{Ribeiro2016Why-arxiv} explains the predictions of any machine learning classifier by approximating any classifier locally with an interpretable model. 
Meanwhile, SHAP (SHapley Additive exPlanations)~\cite{Lundberg2017AUnified-arxiv} offers a unified perspective on feature importance by averaging all feature combinations, ensuring consistent attributions.
%Integrated Gradients attribute deep network predictions to input features by integrating the network’s gradients with respect to its input along a straight-line path from a baseline input to the actual input \cite{Sundararajan2017Axiomatic-arxiv}.
Grad-CAM (Gradient-weighted Class Activation Mapping)~\cite{Selvaraju2016GradCAM-arxiv} employs gradient information from the final convolutional layer to generate a heat map highlighting crucial input image regions.
Counterfactual Explanations~\cite{Wachter2017Counterfactual-arxiv} identify the least number of changes required in the input data to alter the model's prediction, offering insights into decision boundaries and feature importance. 
Lastly, Explanation via Language~\cite{Lakkaraju2022Rethinking-arxiv} emphasizes natural language dialogues for enhanced interaction between experts and models, underscoring the importance of interactive systems tailored to user requirements. 
In this paper, we investigate how large language models can explain decisions, using football refereeing as a testing environment given its decision complexity and subjectivity.

\begin{table*}[t]
    \centering
    \resizebox{\linewidth}{!}{% <------ Don't forget this %
    \begin{tabular}{l|cccc|ccc}
    \toprule
\textbf{Dataset} & \textbf{Type} & \textbf{\#Instances} & \textbf{\#Questions}
& \textbf{\#Context} &\textbf{VQA} &\textbf{Captioning} &\textbf{AR} \\\midrule \rowcolor[HTML]{EFEFEF}

Conceptual 12M~\cite{Changpinyo2021Conceptual-arxiv} & Images & $12M$ & $12M$ & Various & - & \checkmark  & - \\

LAION-5B~\cite{Schuhmann2022LAION5B-arxiv} & Images & $3B$ & $3B$ & Various  & - & \checkmark  & - \\\rowcolor[HTML]{EFEFEF}

LLava dataset \cite{Liu2023Visual-arxiv}  & Images & $158k$ &$158k$  & Various   & \checkmark & \checkmark   & - \\

MovieQA \cite{Tapaswi2015MovieQA-arxiv} & Videos & $408$ & $15k$ & Movies  & \checkmark & - & -  \\\rowcolor[HTML]{EFEFEF}

TVQA \cite{Lei2018TVQA-arxiv} & Videos & $21k$ &  $150k$  & TV shows & \checkmark & -  & - \\

Video Instruction Dataset~\cite{Maaz2023VideoChatGPT-arxiv} &Videos & $100k$ & $100k$ & Various    & \checkmark & \checkmark & - \\\rowcolor[HTML]{EFEFEF}

HowTo100M \cite{Miech2019HowTo100M-arxiv}  & Videos & $136M$ &$136M$ & Youtube  & -  & \checkmark   & -\\\midrule

GOAL \cite{Suglia2022Going-arxiv}  &Videos &$1k$  &$53k$ & Football  & - & \checkmark  & - \\\rowcolor[HTML]{EFEFEF}

Sports-QA \cite{Li2024SportsQA-arxiv}  &Videos &$6k$ &$94k$ &Football  & \checkmark  & -  & \checkmark \\

SoccerNet-caption~\cite{Mkhallati2023SoccerNetCaption}  &  Videos & $942$ &  - & Football games  & - & \checkmark & - \\  \midrule

\textbf{SoccerNet-XFoul (Ours)} & Football &$10k$ & $22k$ & Football fouls    & \checkmark & \checkmark & \checkmark \\ 
\bottomrule
    \end{tabular}
    }
    \caption{ \textbf{Comparative overview of relevant datasets.} \textit{SoccerNet-XFoul} contains high-quality answers annotated by more than 70 experienced referees. Our dataset is the largest dataset in sports with complex questions and the only one focusing on refereeing questions. 
    VQA stands for Visual Question Answering, AR stands for Action Recognition.
    }
    \label{tab:MOT_Comparison}
\end{table*}

\section{SoccerNet-XFoul dataset}
\label{sec:dataset}

The performance of supervised models mostly relies on the quality and quantity of available annotated datasets. Multi-modal datasets are generally harder to curate and annotate, which explains their usually smaller size compared to uncurated datasets.
Table~\ref{tab:MOT_Comparison} shows a comparative overview of multi-modal datasets in the literature, specifically highlighting those that contain combinations of text and image, as well as text and video pairs.
We introduce \emph{SoccerNet-XFoul}, a dataset specifically designed for foul video recognition and explanation. 
It consists of high-quality video-text pairs with more than $10$k video clips and $22$k questions, annotated by more than $70$ experienced referees. 
Compared to the other sports datasets, \textit{SoccerNet-XFoul} has the most video clips and much more complex questions. 
In the following, we present our approach to creating this high-quality human-annotated dataset.

% It is common to resort to automated pipelines that leverage deep learning models for classification, detection, visual question answering, and text augmentation such as rephrasing~\cite{Liu2023Visual-arxiv, Maaz2023VideoChatGPT-arxiv, Li2024SportsQA-arxiv}. 

% such as language models, object-detection, or classification models, to automatically generate question-answer pairs or to augment existing question-answer pairs

% Sports-QA \cite{Li2024SportsQA-arxiv} has generated most of its questions automatically, and the answers are mostly single words 29\% of the answers are either 'yes' or 'no').

% \input{figures/dataset_examples}

\mysection{Questions.}
We identify $4$ key questions on the most foundational, complex, and game-impacting decisions a referee must confront during a game.
To answer the two first questions,
\textit{``Is it a foul or not? Why?''} and \textit{``What card would you give? Why?''}, the model requires an in-depth understanding of the rules of the game~\cite{IFAB2022Laws} as well as an understanding of the context in which an action occurred. Factors such as the intent, the foul location, the game dynamic and the intensity of the contact must all be considered.
The two last questions, if \textit{``the defender stops a promising attack or a goal-scoring opportunity?''} and if \textit{``the referee could have given an advantage?''} add a new layer of difficulty and prediction analysis. The answers to the questions are not only visual, since the model has to make predictions about potential future outcomes. 
For instance, in assessing whether the referee should have given an advantage, the model needs to evaluate whether the attacking team would benefit more from continuing play rather than being granted a free-kick.
% For example, to determine if the referee should have given advantage, the model has to evaluate the likelihood that the attacking team would benefit more from continuing to play than from being awarded a free-kick.

\mysection{Annotators.} 
As no public dataset is available that provides detailed answers and explanations to these refereeing questions, we conducted an annotation campaign with over $70$ referees over a three-month period.
To ensure high-quality answers, only experienced referees were selected for the annotations. 
The referees have officiated between $140$ and $2{,}279$ official games, with an average of $655$ games.
%On average, these referees have officiated $655$ official games each, ranging from $140$ to $2279$ official games.
%from 10 $\sigma=416$).
They were allowed to assess as many video clips as they wished, with the flexibility to pause at any time to avoid fatigue.
Each annotator had the option to provide answers in German, French, English, or Spanish to prevent any linguistic barriers.
The answers were translated from the original language to English by ChatGPT-3.5~\cite{Brown2020Language-arxiv} and then reviewed by another human referee to ensure accurate translation.
% Figure~\ref{fig:dataset_example} shows two annotations from our dataset.

\mysection{Subjectivity.}
Figure~\ref{fig:dataset_example} displays an example of the subjectivity of the annotations. While one referee annotator would not give a card because he thought the foul was made with low intensity, the other annotator would give a yellow card because he believed the tackling was made with high speed and was reckless.
Due to this inherent subjectivity in refereeing, our objective was to gather multiple answers for the same action, rather than collecting a single decision and explanation for each question.
Therefore, the multiple decisions and explanations actually help the model to learn a range of valid interpretations and reasoning strategies employed by human referees. All in all, this can improve the robustness of the AI model, enabling it to make informed decisions even in ambiguous or subjective situations.
Practically, the annotators were randomly assigned different video clips, ensuring that the same action might be evaluated multiple times. In the end, for each action, we have, on average, $1.5$ answers for the same question.

\begin{figure}
  \centering
  \includegraphics[width=1\linewidth]{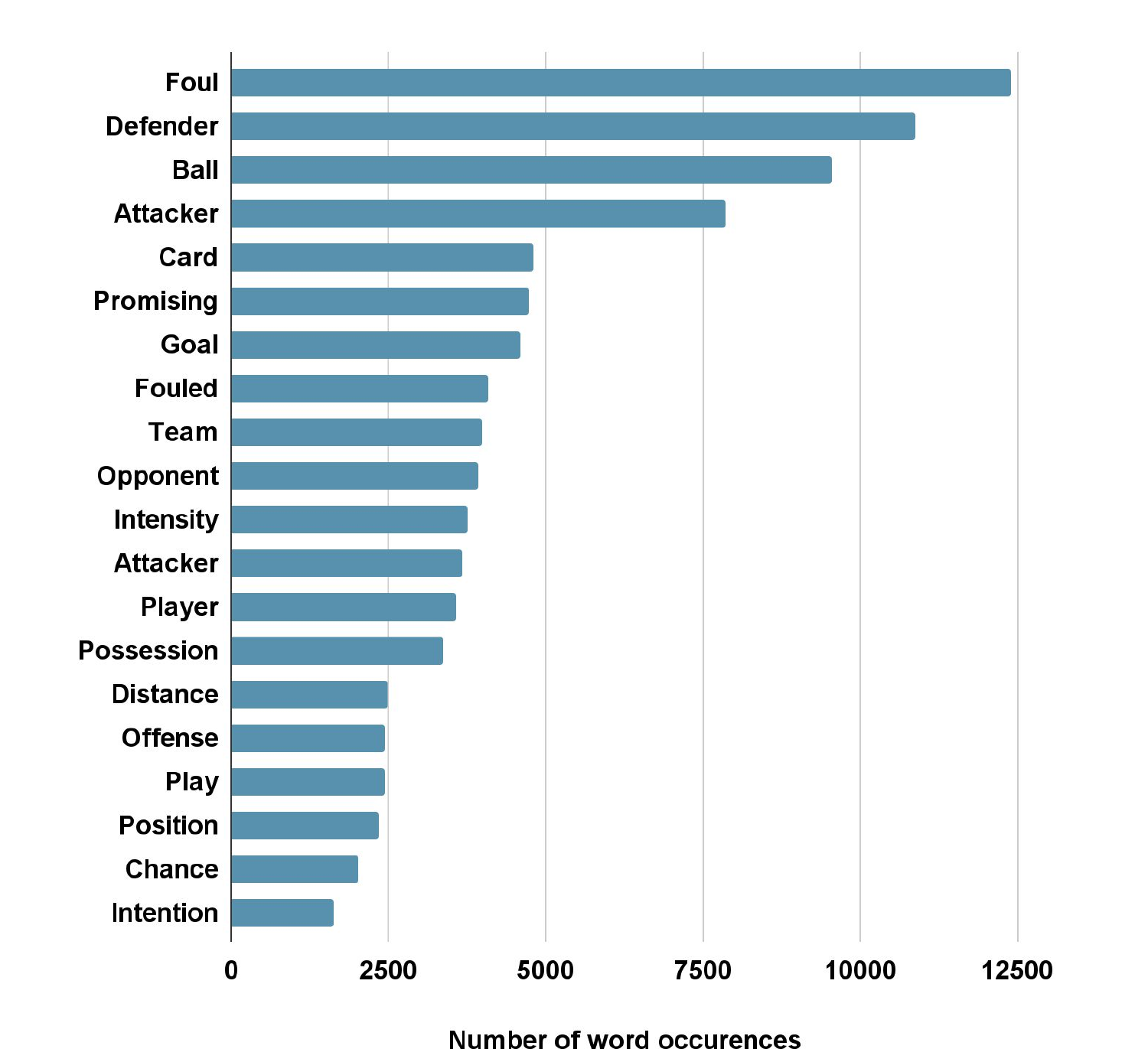}
\caption{
\textbf{Distribution of the most common words.} 
The most frequent words are ``foul" and ``defender," followed by semantically related words related to football and referee actions and terms. There is thus a significant imbalance in the distribution.
}
\label{fig:distribution}
\end{figure}

\mysection{Statistics.} 
Our dataset contains $10$k video clips with over $22$k referee-generated questions and answers.
Figure~\ref{fig:distribution}, shows the distribution of the most common words in the explanations of the referee annotators. 
The most frequently used words are specific terminologies for describing a duel between two players, ranging from descriptive terms such as \textit{defender} or \textit{card} to key terms to consider for evaluating fouls such as \textit{intention} or \textit{intensity}.
%The dataset contains a significant imbalance in the distribution of words.
The number of words per answer ranges from $1$ to $66$, with a total of more than $540$k words and an average of almost $25$ words per answer, with a significant imbalance in the distribution of words.

\mysection{Novelty.} 
Compared to traditional visual-question answering datasets, the \textit{SoccerNet-XFoul} dataset is the first to answer refereeing questions, with detailed explanations of why a particular decision is correct. 
% based on the Laws of the Game~\cite{IFAB2022Laws}.
This explainability is a unique approach that enhances the dataset’s complexity and ensures a deeper understanding and representation of real-world scenarios where AI models must make and explain their decisions.
Furthermore, the interpretation of situations in our dataset is context-dependent. 
The interpretation of a foul might differ depending on whether it occurs in the middle of the field or in the penalty area. To correctly answer questions, the model must have an in-depth understanding of the game.
Finally, the model needs a level of predictive analysis. For instance, to determine if a defender stopped a promising attack, the model must understand what happened at the moment of the foul and what could have happened in the immediate future. This involves making complex predictions about potential future outcomes, a task that is far more advanced compared to traditional VQA datasets.
Hence, our \textit{SoccerNet-XFoul} dataset is the first and largest visual question-answering dataset for referees in football, offering many new challenges to be explored.

%Additionally, since a part of our dataset comprises multiple views, the model needs to combine information from different perspectives to arrive at the correct decision. Not all answers are visible from each view. Often, each view contains hints that need to be combined to arrive at the correct decision. To determine if a defender stopped a promising attack, the model must first determine if there was a foul (often determined in a close-up view), while the main view provides indications of where the foul occurred and whether it prevented a promising attack. 
%As far as our knowledge, this task has not yet been explored for visual question answering.

%\mysection{Metrics.}
%Traditional metrics like METEOR, BLEU, ROUGE, and CIDEr are not applicable to measure the performance of our model.
%For instance, two texts "The referee gave a yellow card" and "The referee gave a red card" might appear linguistically similar, yet in the context of refereeing, the meaning of each is very different.
%To assess the quality of the model's referee decisions and explanations qualitatively, we conduct a human study. In this study, 15 experienced referees evaluate the generated texts against three criteria: the overall quality of the text, the accuracy of the decision, and the clarity of the decision's explanation.

\begin{figure*}[t]
  \centering
  \includegraphics[width=0.8\linewidth]{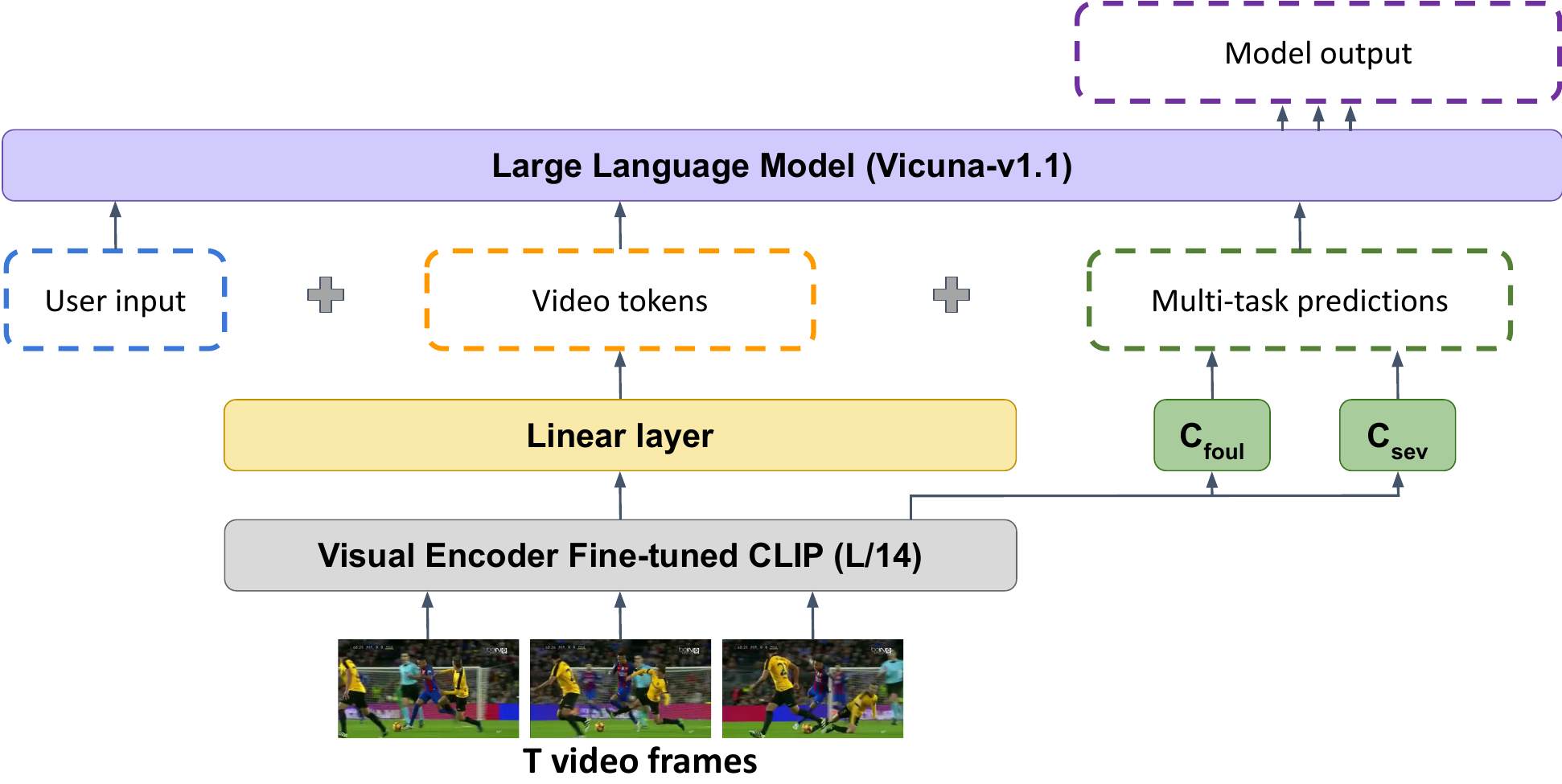}
\caption{
\textbf{Architecture of X-VARS.}  
X-VARS is a visual language model based on a fine-tuned CLIP visual encoder to extract spatio-temporal video features and to obtain multi-task predictions regarding the type and severity of fouls. The linear layer connects the vision encoder to the language model by projection the video features in the text embedding dimension. We input the projected spatio-temporal features alongside the text predictions obtained by the two classification heads $\mathbf{C_{foul}}$ and $\mathbf{C_{\severity}}$ (for the task of determining the type of foul and the task of determining if it is a foul and the corresponding severity) into the Vicuna-v1.1 model, initialized with weights from LLaVA.
}
\label{fig:architecture}
\end{figure*}

\section{Methodology}
\label{sec:method}
In this section, we provide a comprehensive description of our novel E\textbf{\underline{X}}plainable \textbf{\underline{V}}ideo \textbf{\underline{A}}ssistant \textbf{\underline{R}}eferee \textbf{\underline{S}}ystem, ``X-VARS'', for foul and severity recognition, providing explanations on its decision-making process. We begin by presenting the architecture with a detailed description of each individual component of X-VARS. Then, we provide an in-depth explanation of its training process.

\subsection{Architecture}

Figure~\ref{fig:architecture} illustrates the key architectural components of X-VARS. We use Video-ChatGPT~\cite{Maaz2023VideoChatGPT-arxiv}, a multi-modal model capable of understanding and generating detailed conversations about videos, as our foundation model. We make several changes to the architecture to adapt it to our needs.
Formally, we input a video clip $\mathbf{v} \in \mathbb{R}^{T\times H \times W \times C}$, with $T$, $H$, $W$ and $C$ being respectively the number of frames, height, width, and channel dimension of the video, to CLIP ViT-L/14~\cite{Radford2021Learning},
\begin{equation}
    \boldsymbol{f_i} , \boldsymbol{h_i} = CLIP(v) \comma
\end{equation}
and obtain the corresponding frame feature vector $\boldsymbol{f_i}$ $\in \mathbb{R}^{T\times D_1}$ and the hidden states $\boldsymbol{h_i}$ $\in \mathbb{R}^{T \times S \times D_2}$, with $S$ being the number of tokens obtained by multiplying $w = W/p$ and $h = H/p$, where $p$ is the patch size of CLIP, $D_1$ the dimension of the output layer and $D_2$ the dimension of the hidden states.
We then average-pool the hidden states across the temporal dimension to obtain temporal features $\textbf{t}\in \mathbb{R}^{S\times D_2}$ and along the spatial dimension to get the video-level spatial representation $\textbf{s}\in \mathbb{R}^{T\times D_2}$.
Finally, we concatenate both to obtain spatio-temporal features.
\begin{equation}
    \boldsymbol{z} = [\boldsymbol{t} \quad \boldsymbol{s}] \in \mathbb{R}^{(S + T) \times D_2} \dotmath
\end{equation}
Before feeding the video features $\boldsymbol{z}$ into the LLM, we project them into the same feature space as the text embeddings by applying a linear projection layer: 
\begin{equation}
    \boldsymbol{w} = Linear(\boldsymbol{z}) \in \mathbb{R}^{(S + T) \times D_2}  \dotmath
\end{equation}
with $\boldsymbol{w}$ being a sequence of visual tokens.
%
% \newline
The feature vectors $\boldsymbol{f_i}$ are also average-pooled along the temporal dimension to obtain a single video-level representation $\boldsymbol{f}$ $\in \mathbb{R}^{D_1}$.
The video-level feature representation $\boldsymbol{f}$ is passed through two classification heads  $\mathbf{C_{foul}}$ and  $\mathbf{C_{\severity}}$ to obtain the type of foul (\ie Tackling, Holding, Pushing, Standing tackling, Elbowing, Dive, Challenge, or High leg) and to determine whether it is a foul or not, and the corresponding severity (\ie No offence, Offence + No card, Offence + Yellow card or Offence + Red card), with the predictions being:
\begin{equation}
    \mathbf{P_{foul}} = \arg\max \mathbf{C_{foul}}\comma
\end{equation}
\begin{equation}
    \mathbf{P_{\severity}} = \arg\max \mathbf{C_{\severity}}\dotmath
\end{equation}
These predicted labels with the highest confidence are injected as a textual prompt to the LLM.
Hence, this multi-task classification enables the model to utilize acquired information to enhance the performance of the explanation.

To obtain high performance with LLMs, a crucial part consists in determining a prompt which is understandable by the LLM. 
As we use the Video-ChatGPT backbone, we design our prompt with the following query:
\begin{flalign}
&USER: <Question> <P_{foul}><P_{\severity}><w>\\\nonumber
&Assistant:&&
\end{flalign}
where $<Question>$ represents one of our questions randomly sampled from the training set of video-question-answer triplets, 
$<P_{foul}>$ and $ <P_{\severity}>$ are the two predictions on the foul type and severity recognition task obtained from the fine-tuned CLIP, and 
$<w>$ are the projected spatio-temporal features.
X-VARS is trained to predict the answers of the assistant as an auto-regressive model.
% X-VARS is trained to predict the assistant's answers and determine when to stop. Only the tokens generated for the assistant are used to compute the loss in the auto-regressive model.

\subsection{Training Paradigm}

%\begin{table}
%\centering
%\resizebox{\linewidth}{!}{%
%\setlength{\extrarowheight}{4pt}
%\begin{tabular}{c|ccccc|}
%  & CLIP &$\mathbf{C^{foul}}$ \&  $\mathbf{C^{\severity}}$ & Linear Layer & LLM \\
%\hline
%Stage 1 & \cincludegraphics[height=.2in]{pictures/fire_real.png} &\cincludegraphics[height=.2in]{pictures/fire_real.png} & - & -\\
%Stage 2 & \cincludegraphics[height=.2in]{pictures/ice_real.png} & \cincludegraphics[height=.2in]{pictures/ice_real.png} & \cincludegraphics[height=.21in]{pictures/fire_real.png} & \cincludegraphics[height=.21in]{pictures/fier_ice.png}\\

%\end{tabular}
%}
%\caption{ \textbf{Overview of the training stages.} In stage 1, we fine-tune CLIP and the classification heads $\mathbf{C^{foul}}$ and $\mathbf{C^{\severity}}$ to give X-VARS some prior nowledge about soccer and refereeing. 
%In stage 2, we keep them frozen and fine-tune the linear layer and partially fine-tune the LLM.
%}
%    \label{tab:fine_tune_stages}

\begin{table}
    \centering
    \resizebox{\linewidth}{!}{% <------ Don't forget this %
    \begin{tabular}{l|lccc}
          &  CLIP &$\mathbf{C_{foul}}$ \&  $\mathbf{C_{\severity}}$ & Linear Layer & LLM  \\ \midrule
Stage 1 & \cincludegraphics[height=.2in]{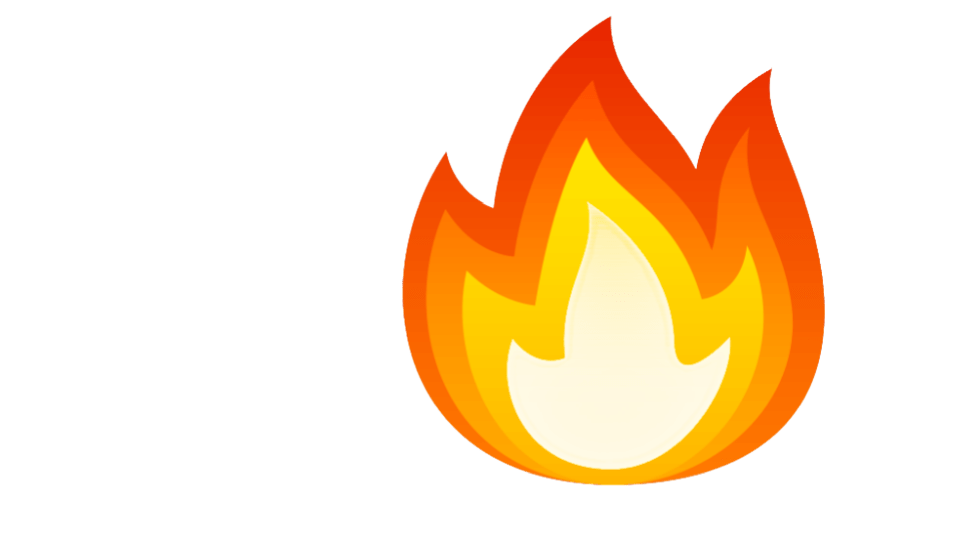} &\cincludegraphics[height=.2in]{pictures/fire_real.png} & - & -\\
Stage 2 & \cincludegraphics[height=.2in]{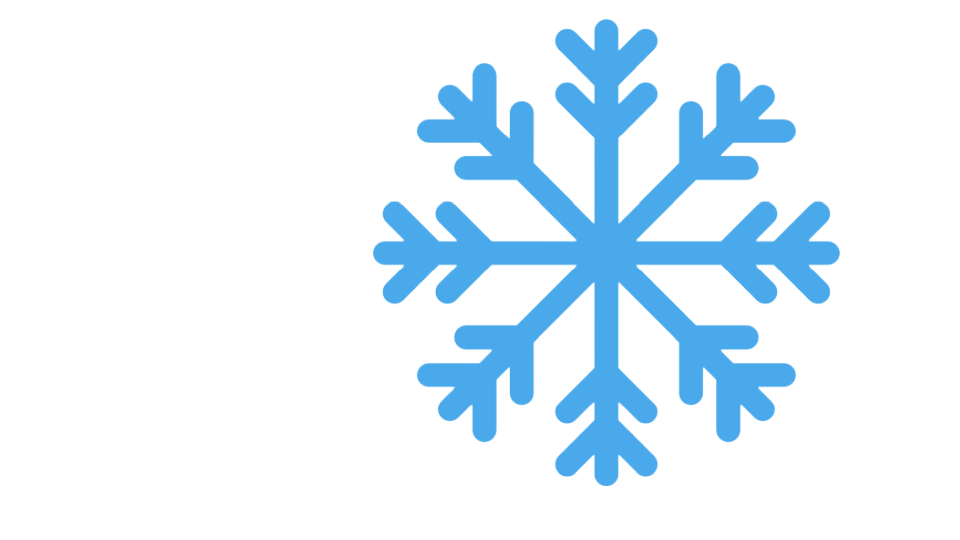} & \cincludegraphics[height=.2in]{pictures/ice_real.png} & \cincludegraphics[height=.21in]{pictures/fire_real.png} & \cincludegraphics[height=.21in]{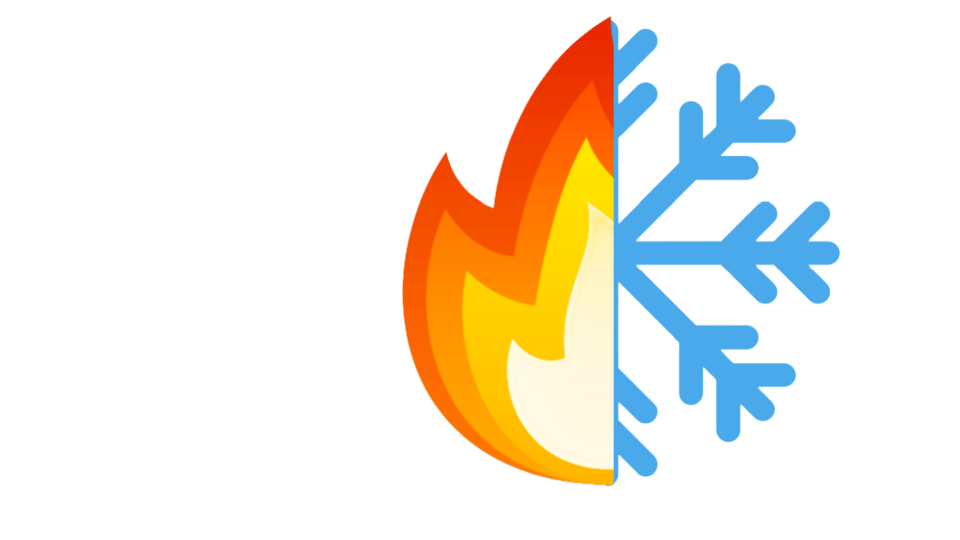}\\
    \end{tabular}
    }
\caption{ \textbf{Overview of the training stages.} In stage 1, we fine-tune CLIP and the classification heads $\mathbf{C_{foul}}$ and $\mathbf{C_{\severity}}$ to give X-VARS some prior knowledge about refereeing. 
In stage 2, we keep them frozen and fine-tune the linear layer and partially the LLM.
}
    \label{tab:fine_tune_stages}
\end{table}

We propose a two-stage training approach. The first stage fine-tunes CLIP on a multi-task classification to learn prior knowledge about football and refereeing. The second step consists in fine-tuning the projection layer and several layers of the LLM to enhance the model's generation abilities in the sport-specific domain.

\mysection{Stage 1: Fine-tuning to inject football knowledge.}
While CLIP is excellent at generalizing across various image tasks, it lacks the ability to recognize fine-grained actions or events. These actions are mostly recognizable by considering the temporal dimension rather than images alone. For instance, assessing the severity of a foul requires considering factors such as the intensity and the speed, which cannot be determined by simply examining images. 
% Since CLIP was not trained on football-specific tasks, the output space projection remains too similar for two football clips, even if the video contents are different. 
Since CLIP was not trained specifically on football data, the feature representation between two football clips would be very similar, despite the videos depicting different scenarios.
Hence, since all our videos are related to football, the output features will share similarities. This proximity between features actually poses a challenge for the LLM, making it difficult to effectively distinguish between different actions.
To avoid these issues, we fine-tune CLIP on the \textit{SoccerNet-MVFoul} dataset~\cite{Held2023VARS} to learn prior knowledge about football. 
The training consists of minimizing the summed cross-entropy loss of both tasks. 
Given the similar magnitudes of both losses, we sum them without scaling or weighting.
% After fine-tuning CLIP, we keep it frozen for the second fine-tuning step.

\mysection{Stage 2: Feature alignment and end-to-end training.}
We freeze the fine-tuned CLIPs weights and continue training the linear projection layer and the LLM.
Training a projection layer from scratch requires many high-quality video-text pairs and computational resources.
To alleviate this, we use the pre-trained weights of the projection layer from Video-ChatGPT~\cite{Maaz2023VideoChatGPT-arxiv}, which was trained on a dataset of $100$k video-text pairs.
We further fine-tune this projection layer to map the spatio-temporal features of our football clips into the same dimensional space as the word embeddings.
As demonstrated in~\cite{Maaz2023VideoChatGPT-arxiv, Liu2023Visual-arxiv, Zhu2023MiniGPT4-arxiv, Lyu2023MacawLLM-arxiv}, a simple linear layer is sufficient for connecting video features with the word embedding.
%To enhance performance, we also fine-tune some layers of the LLM.
During training, we replace the predictions of CLIP $<P_{foul}>$ and $<P_{\severity}>$ by the ground truth $<G_{foul}>$ and $ <G_{\severity}>$ as CLIPs predictions might be noisy, which could lead to confusions.
Since determining foul type and severity is subjective, alignment between the ground truth of the \textit{SoccerNet-MVFoul} dataset and the referee responses from our \textit{SoccerNet-XFoul} dataset may vary. Consequently, even when giving $<G_{foul}>$ and $<G_{\severity}>$ during training, the model may not only use this information without using the video tokens to produce the text. % because the ground truth is not always in alignment with the referees answer.

\begin{figure*}
    \centering
    \begin{subfigure}[t]{0.49\textwidth}
        \centering
        \includegraphics[width=\linewidth]{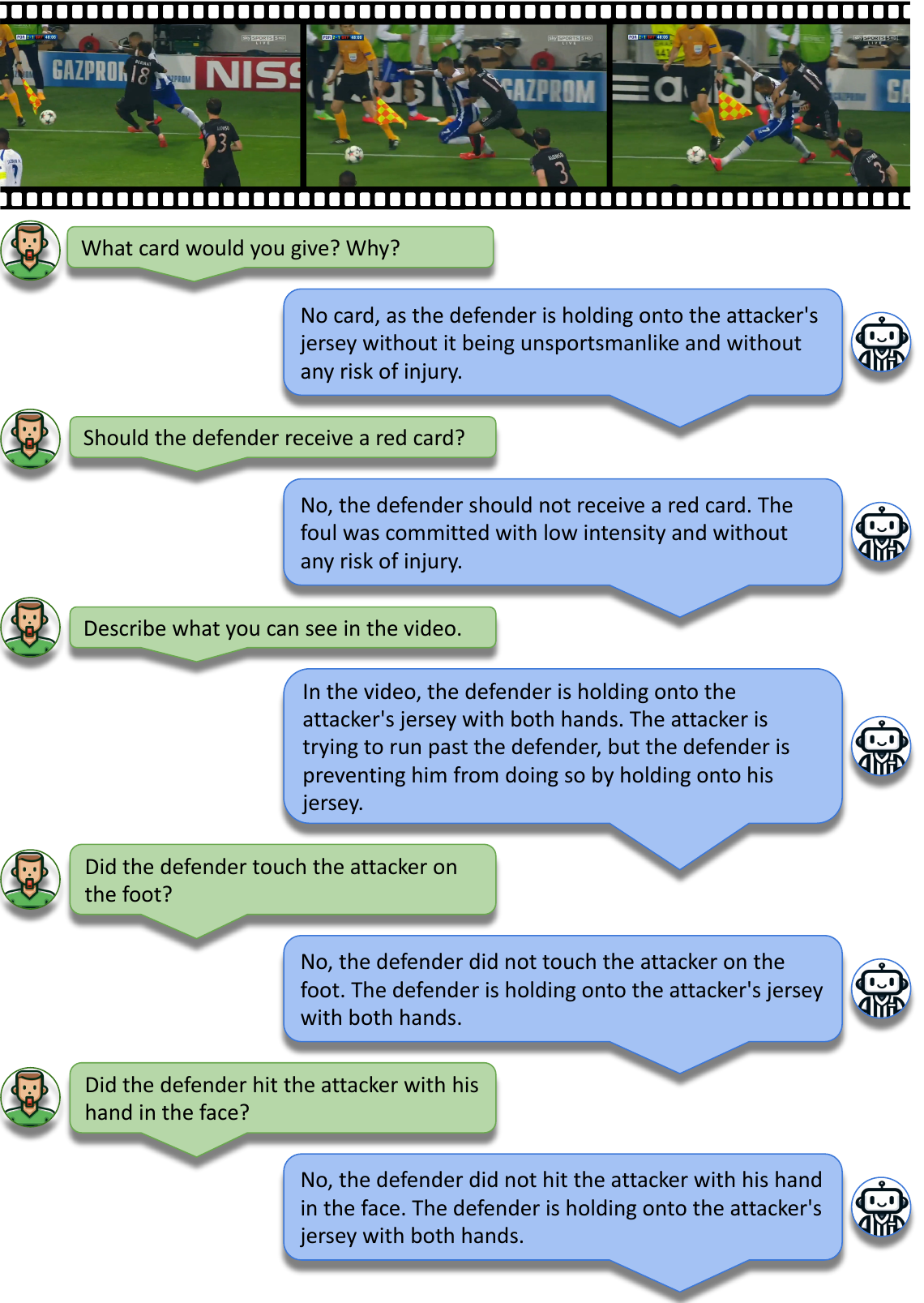}
        \caption{\textbf{Q:} What card would you give? Why? \textbf{GT:} No card because the defender briefly held onto the attacker's arm during the fight for the ball, without it being unsportsmanlike.
        }
        \label{fig:qualitative_res_a}
    \end{subfigure}
    \hfill
    \begin{subfigure}[t]{0.49\textwidth}
        \centering
        \includegraphics[width=\linewidth]{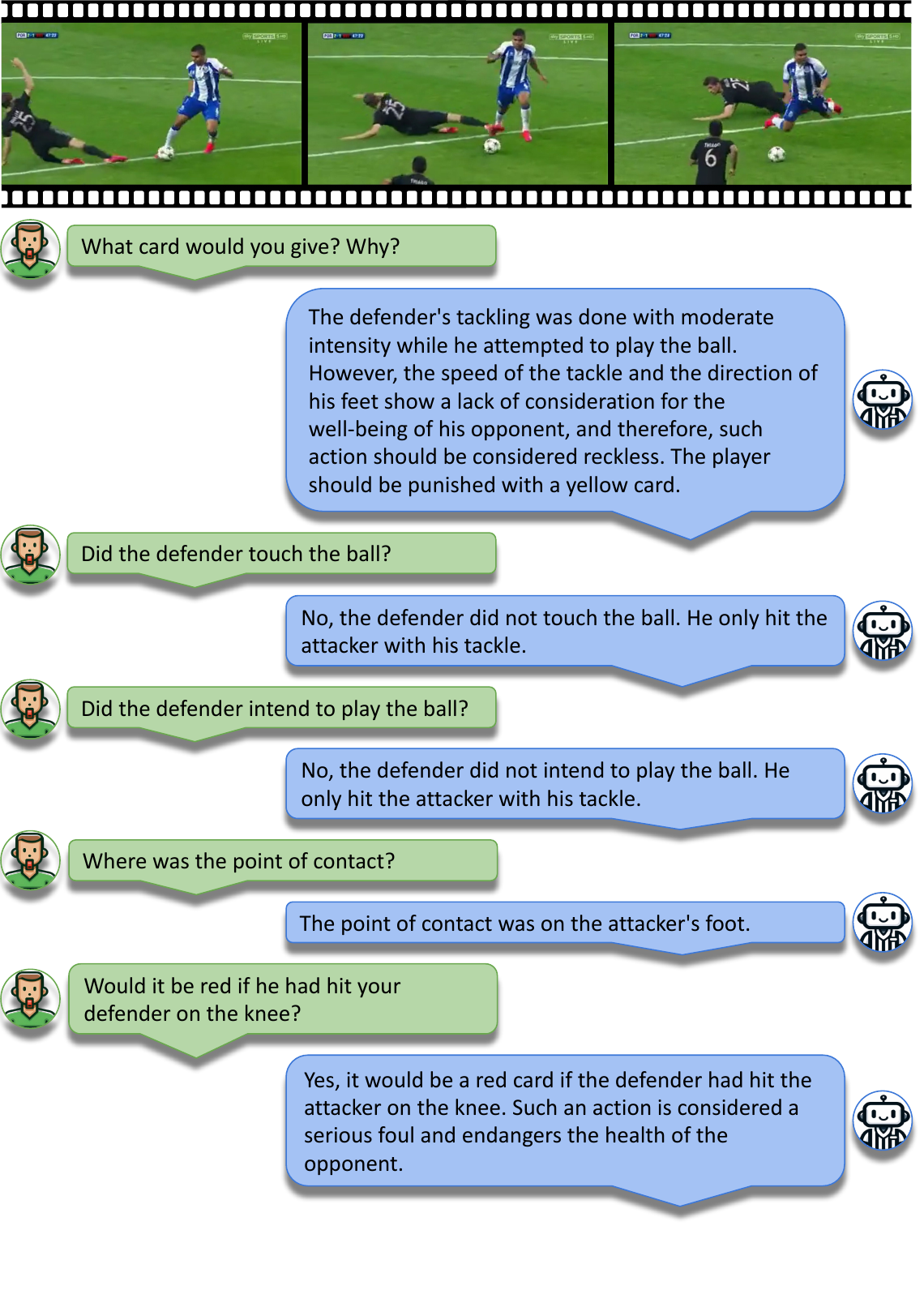}
        \caption{\textbf{Q:} What card would you give? Why? \textbf{GT:}  No card. Even though the defender had no chance to play the ball, he touched the attacker with low intensity on the foot.}
        \label{fig:qualitative_res_b}
    \end{subfigure}
    \caption{
\textbf{Qualitative results.}
Although X-VARS has never been specifically fine-tuned for conversation, it has inherited its conversational capabilities from the pre-trained model. X-VARS demonstrates impressive discussion skills while being aligned with the video content and the Laws of the Game.
(a) X-VARS is close to the ground truth and is able to accurately answer the user's question.
(b) This example shows the subjectivity of foul situations. X-VARS interprets the foul as medium intensity, while the human referee interprets it as low intensity with no chance to play the ball.
}
    \label{fig:qualitative}
\end{figure*}

\section{Experiments} \label{sec:benchmarks}

In this section, we analyze the performance of X-VARS on the two most important refereeing questions: \textit{``Is it a foul or not? Why?''} and \textit{``What card would you give? Why?''}. Given the importance of these questions, we conduct a comprehensive and detailed analysis, providing insights into the improvements in the video recognition performance, a human study to assess the model's explanations, some qualitative results, and a thorough ablation study. 

\subsection{Implementation details}
We fine-tune CLIP-L/14 on the \textit{SoccerNet-XFoul} dataset for $14$ epochs with a learning rate of $5\times10^{-6}$ on a single Nvidia V100 GPU with a batch size of $64$, using gradient accumulation to 
overcome memory limitation.
% increase the effective batch size during training. 
The fine-tuning of the model takes about $9$ hours. We use $16$ frames in $224$p resolution per clip, with $8$ frames before and $8$ frames after the foul. %The quality of the frames is 224p.
% For the second stage, we keep CLIP-L/14 frozen and only train the linear projection layer and some layers of the LLM. 
% Table~\ref{tab:fine_tune_stages} provides an overview of the state of the various key components during training.
For the second stage, we employ QLORA~\cite{Hu2021LoRA-arxiv, Dettmers2023QLoRA-arxiv} to enhance memory efficiency and enable training on a single GPU.
%QLORA effectively reduces the memory usage of the model while maintaining its training capability. 
%By implementing QLORA in our model architecture, we were able to mitigate memory constraints and leverage the computational power of a single GPU for training.
We only fine-tune $1$\% of the layers for $3$ epochs using a learning rate of $2\times10^{-4}$ and an
overall batch size of $32$. The training on $2$ A100 $40$GB GPUs takes about $2$ hours.
Table~\ref{tab:fine_tune_stages} provides an overview of the state of the various key components during training.

\begin{table}
    \centering
    \resizebox{\linewidth}{!}{% <------ Don't forget this %
    \begin{tabular}{l|c|lcccc}
        \multicolumn{1}{c|}{} &  \multicolumn{1}{c|}{} & 
        \multicolumn{5}{c}{\bf Distribution} \\ \midrule
        & \bf Mean  & \bf 1 & \bf 2 & \bf 3 & \bf 4 & \bf 5\\ \midrule
       Referees  & 4.0     &3\% &10\% &8\% &46\% &33\% \\ 
       X-VARS  & 3.8  &3\% &17\% &4\% &46\% &30\%  \\ 
    \end{tabular}
    }
%\caption{ \textbf{Quantitative results.} 
\caption{ \textbf{Score and distributions obtained during our human study comparing the quality of referees and X-VARS generated explanations.}
The mean scores of X-VARS closely match those of human referees. In 46\% of the video clips,
X-VARS achieved higher scores for its explanations than the human referees. 
The distribution of the results is very similar for human referees and X-VARS. 
A score of 5 is the highest and represents \emph{strongly agree} while 1 is \emph{strongly disagree}.
}
    \label{tab:quantitative_distribution}
\end{table}
\subsection{Human study on explanation performance}
Evaluating generative tasks, such as text, image, or video generation, remains a significant challenge due to their subjective nature and the absence of proper evaluation metrics. 
Traditional language metrics are not very informative for our purpose, as two sentences can be linguistically very similar, yet have entirely different meanings. 
To achieve quantitative results, we conducted a human study with $20$ football referees who evaluated the quality of responses without knowing if they were generated by a human referee or by X-VARS. 
The referee officiated between $85$ and $850$ official games, with an average of $490$ games.
%On average, these referees have officiated 483 official games ($\sigma=214$ official games) and were not part of the annotation campaign.
Each participant assessed $20$ random video clips, each lasting $5$ seconds, with no time restrictions. 
They evaluated the quality of the explanation, considering whether the evaluation was consistent with the video and if the decision and explanation aligned with the Laws of the Game~\cite{IFAB2022Laws}.
They rated each explanation on a scale of $1$ to $5$, with $5$ indicating \textit{strongly agree} and $1$ indicating \textit{strongly disagree}.
Table~\ref{tab:quantitative_distribution} shows the results, with X-VARS performing similarly to the human referees, with only minimal score differences. 
X-VARS's explanations were more convincing in $46\%$ of the cases than the referee's. 
While both show similar results for \textit{strongly agree} and \textit{agree}, X-VARS obtains more \textit{disagree} responses than human referees. 
The majority of videos where participants disagreed with X-VARS involve types of fouls that are rare in our dataset, \ie when the defender uses his arms illegally by pushing his opponent or hitting him with the elbow in the face.
Overall, the human study highlights X-VARS's impressive ability to understand football videos and explain its decisions at a level comparable to human referees.

\subsection{Qualitative results}

Figure~\ref{fig:qualitative} showcases two examples of conversations generated by our proposed X-VARS. Particularly, we illustrate its remarkable ability to understand and generate decisions with explanations related to visual content and the Laws of the Game~\cite{IFAB2022Laws}. 
Although X-VARS was only fine-tuned on two questions, Figure~\ref{fig:qualitative} illustrates that X-VARS can generalize and accurately answer or describe video content without any specific fine-tuning. 
Furthermore, X-VARS was not fine-tuned for conversation, but we inherited these capabilities from the pre-trained conversational model Video-ChatGPT~\cite{Maaz2023VideoChatGPT-arxiv}, which serves as the foundation for X-VARS. 
Hence, throughout our two fine-tuning stages, we retained the conversation capabilities of the foundation model and can generate meaningful conversations with X-VARS about football and refereeing.
Another interesting fact is the typical characteristic of LLMs to consistently agree with human users. 
Surprisingly, X-VARS mostly maintains its decision and offers comprehensive explanations for it, even when asked questions such as \textit{``Should the defender receive a red card?''}, when the specific foul would not require any.
However, similarly to other LLMs, X-VARS has also inherited typical issues such as hallucinations, in which it recognizes actions in the video that are not present. 
Future work could investigate if more high-quality data or more advanced LLMs would limit this hallucination effect.

\subsection{Ablation study}

\begin{table}[t]
    \centering
    \resizebox{\linewidth}{!}{% <------ Don't forget this %
    \begin{tabular}{lc|cc|lc}
        \multicolumn{2}{c|}{} &  \multicolumn{2}{c|}{\bf Type of Foul} & 
        \multicolumn{2}{c}{\bf Offence Severity} \\ \midrule
       \bf Feat. extr. & \bf Pooling  & \bf Acc. & \bf BA. & \bf  Acc. & \bf BA.\\ \midrule
       ResNet~\cite{Held2023VARS} & Mean     &0.30     & 0.27  &0.34     &0.25 \\ 
       ResNet~\cite{Held2023VARS}  & Max      &0.32  &0.27   & 0.32  &0.24 \\
       R(2+1)D~\cite{Held2023VARS} & Mean     &0.32     & 0.34  &0.34     &0.30 \\ 
       R(2+1)D~\cite{Held2023VARS}  & Max      &0.34  &0.33   & 0.39  &0.31 \\
       VARS-MViT~\cite{Held2023VARS}  & Mean     &0.44     & \textbf{0.40}   &0.38     &0.31 \\ 
       VARS-MViT~\cite{Held2023VARS}  & Max      &0.45  &0.39   & 0.43  &0.34 \\\midrule
       CLIP-L/14 & Single-view &\textbf{0.51}  & 0.39   &0.52 &\textbf{0.35} \\
       X-VARS & Single-view  & /  &/  &\textbf{0.62} &\textbf{0.35}
    \end{tabular}
    }
    \caption{\textbf{Multi-task classification.} We compare the multi-task classification accuracy of the fine-tuned CLIP-L/14 and X-VARS (fine-tuned CLIP + LLM) to the performance obtained by Held \etal ~\cite{Held2023VARS}. We obtain state-of-the-art performances for three of the four metrics while using a single view instead of multi-views. X-VARS enhances the classification accuracy of offence and severity by 19\% compared to the previous SOTA.
    \textbf{Acc.} stands for Accuracy and \textbf{BA.} stands for balanced accuracy.
    }
    \label{tab:classification}
\end{table}

\mysection{Video action recognition performance.}
Table~\ref{tab:classification} shows the performance of CLIP-L/14 after the fine-tuning process in foul classification tasks. We compare it to the previous state-of-the-art (SOTA) achieved by Held~\etal~\cite{Held2023VARS} by using the same number of frames and video quality to have a fair comparison.
Held~\etal used the MViT~\cite{li2022mvitv2, fan2021multiscale} video encoder to extract spatio-temporal features. As \textit{SoccerNet-MVFoul} contains multiple views for each action, they aggregate the features of the different views by mean or max pooling.
In this work, we fine-tune CLIP on a single view and evaluate it on the same actions, using only a single view instead of the multi-views. 
Despite fewer views, CLIP outperforms the previous SOTA in three of the four metrics, especially enhancing foul and severity classification, with an increase of 9\% in accuracy.
% the performance is increased by 9\%. 
To determine the recognition performance of X-VARS, we asked X-VARS for each video clip, if it was a foul or not, and its corresponding severity.
We then asked ChatGPT-3.5 to extract the classification predictions from the generated explanations. 
Finally, comparing these predictions of X-VARS to the ground truth, we observe a significant performance increase to $62\%$ accuracy in determining whether a foul occurred and its severity compared to CLIPs predictions.
%, corresponding to an increase of $10\%$. 
Hence, X-VARS outperforms the previous state-of-the-art VARS (MViT+Max Pool) system~\cite{Held2023VARS} by $19\%$. 
However, since most of the explanations of X-VARS do not explicitly indicate the type of foul, it is not possible to accurately extract it from the explanations.
For this reason, we were not able to evaluate the accuracy of X-VARS in determining the type of foul.
%Figure~\ref{fig:dataset_example} shows an example of where it is impossible to extract the accurate type of action from the explanations.

\mysection{Influence of the video tokens.}
%\mysection{CLS improvement/refinement on top of the cls prediction.}
% \mysection{Influence of CLIP's classification predictions.}
% An important ablation study involves analyzing the influence of the prediction input in our language model. 
We investigated if the LLM simply generates its answers based on the multi-task predictions that we give as input to the LLM or if it also considers the video tokens.
To test this hypothesis, we use the X-VARS prediction obtained in the previous section, and we compare it to the CLIPs prediction provided as input. Interestingly, X-VARS did not simply replicate the CLIP prediction as it only agreed on $76\%$ of the cases.
This result shows that X-VARS, throughout its training, developed the ability to re-evaluate the multi-task predictions and understand that they are not always reliable.
Consequently, X-VARS does not only rely on the text predictions for its answers but also incorporates information from the video tokens.

\mysection{CLIPs classification predictions \vs no text predictions.} 
To validate our new training paradigm, we compared the quality of our X-VARS trained with classification prediction as additional text input against training it only with video features. To compare the two models, we randomly selected a set of $40$ video clips with a uniform distribution of various types of actions and severities. % (where possible. For instance, only 2 red cards are available in the test set).
By qualitatively analyzing the results on the selected set, both models generate similar outcomes for most of the video clips. The main difference occurs for less frequent types of actions and severities. For instance, X-VARS without predictions fails to predict a single ``No foul'' instance and achieves a balanced accuracy of only $29\%$, while X-VARS obtains a $6\%$ higher balanced accuracy. 
Figure~\ref{fig:qualitative_res_a} shows a clip of a defender holding his opponent, an underrepresented action in our dataset. X-VARS without predictions incorrectly predicts: ``\textit{No card, as the defender pushed the attacker in the back with low intensity during the fight for the ball, without any risk of injury}''.
On the other hand, X-VARS with predictions provides an accurate explanation: ``\textit{No card, as the defender is holding onto the attacker’s jersey without it being unsportsmanlike and without any risk of injury.}''. 
Throughout our testing, we encountered several instances where X-VARS with prediction tokens aligned more closely with the ground truth, especially for underrepresented actions.
These results show the effectiveness of our new training paradigm in achieving higher accuracy and more accurate explanations.

\section{Conclusion}
\label{sec:conclusion}
In this work, we investigated the potential of using LLMs to enhance transparency and explainability within decision-making processes.
We proposed X-VARS, a multi-modal language model, which can perform a multitude of tasks, including video description, question answering, video action recognition, and conducting meaningful conversations based on video content. 
%apply the Laws of the Game, engage in accurate conversations about soccer video clips, and generate captions aligned with the visual content.
X-VARS achieves state-of-the-art performance in determining whether a duel between players constitutes a foul and in assessing the corresponding severity. 
The qualitative results and human study underscore the exceptional capabilities of X-VARS in explaining its decision, indicating its potential to enhance football refereeing by providing accurate decisions and explanations.
{
% \tiny
% \scriptsize
% \footnotesize
% \small
% \normalsize

\mysection{Acknowledgement.}
This work was partly supported by the King Abdullah University of Science and Technology (KAUST) Office of Sponsored Research through the Visual Computing Center (VCC) funding and the SDAIA-KAUST Center of Excellence in Data Science and Artificial Intelligence (SDAIA-KAUST AI). 
J. Held and A. Cioppa are funded by the F.R.S.-FNRS. The present research benefited from computational resources made available on Lucia, the Tier-1 supercomputer of the Walloon Region, infrastructure funded by the Walloon Region under the grant agreement n°1910247.

}

\cleardoublepage

%%%%%%%%% REFERENCES
{\small
\bibliographystyle{ieee_fullname}
\bibliography{bib/abbreviation-short,
bib/activity,
bib/action,
bib/learning,
bib/dataset,
bib/labo,
bib/llm,
bib/soccer,
bib/soccernet-challenge,
bib/sports,
bib/explainability}
}

%bib/put-new-refs-here,
%bib/mllm,
%bib/bib,

\cleardoublepage

\end{document}